\documentclass[10pt,twocolumn,letterpaper]{article}

\usepackage{style/iccv}

\date{}
\usepackage[english]{babel}

\usepackage{mathtools} 
\usepackage{graphicx}
\usepackage{wrapfig}
\usepackage{multicol}
\usepackage{subfig}
\usepackage{booktabs}
\usepackage{amsfonts}
\usepackage{gensymb}
\usepackage{listings}
\usepackage{multirow}
\usepackage{pifont}

\usepackage[colorinlistoftodos,prependcaption,textsize=tiny]{todonotes}

\usepackage[normalem]{ulem}

\usepackage[bottom]{footmisc}
\usepackage[colorlinks=true, allcolors=blue]{hyperref}

\usepackage{array,tabularx,calc}

\newcommand{\xmark}{\text{\ding{55}}}

\newcommand{\ra}[1]{\renewcommand{\arraystretch}{#1}}

\usepackage{filecontents}

\title{\LARGE \bf
sharpDARTS: Faster and More Accurate\\ Differentiable Architecture Search
}

\author{Andrew Hundt$^1$, Varun Jain$^1$, Gregory D. Hager$^1$\\
Johns Hopkins University Department of Computer Science\\
{\tt\small \{ahundt, vjain, ghager1\}@jhu.edu}}

\iccvfinalcopy 
\def\iccvPaperID{****} 

\begin{document}
\maketitle
\begin{abstract} 
Neural Architecture Search (NAS) has been a source of dramatic improvements in neural network design, with recent results meeting or exceeding the performance of hand-tuned architectures.
However, our understanding of how to represent the search space for neural net architectures and how to search that space efficiently are both still in their infancy. 

We have performed an in-depth analysis to identify limitations in  a widely used search space and a recent architecture search method, Differentiable Architecture Search (DARTS).
These findings led us to introduce novel network blocks with a more general, balanced, and consistent design; a better-optimized Cosine Power Annealing learning rate schedule; and other improvements.
Our resulting sharpDARTS search is 50\% faster with a 20-30\% relative improvement in final model error on CIFAR-10 when compared to DARTS.
Our best single model run has 1.93\% (1.98$\pm$0.07) validation error on CIFAR-10 and 5.5\% error (5.8$\pm$0.3) on the recently released CIFAR-10.1 test set. 
To our knowledge, both are state of the art for models of similar size. 
This model also generalizes competitively to ImageNet at 25.1\% top-1 (7.8\% top-5) error.

We found improvements for existing search spaces but does DARTS generalize to new domains?
We propose Differentiable Hyperparameter Grid Search and the HyperCuboid search space, which are representations designed to leverage DARTS for more general parameter optimization.
Here we find that DARTS fails to generalize when compared against a human's one shot choice of models.
We look back to the DARTS and sharpDARTS search spaces to understand why, and an ablation study reveals an unusual generalization gap.
We finally propose Max-W regularization to solve this problem, which proves significantly better than the handmade design.
Code will be made available.

\end{abstract}
\section{Introduction}

\begin{figure}[bt!]
    \centering
    \hfill
    \includegraphics[width=\columnwidth]{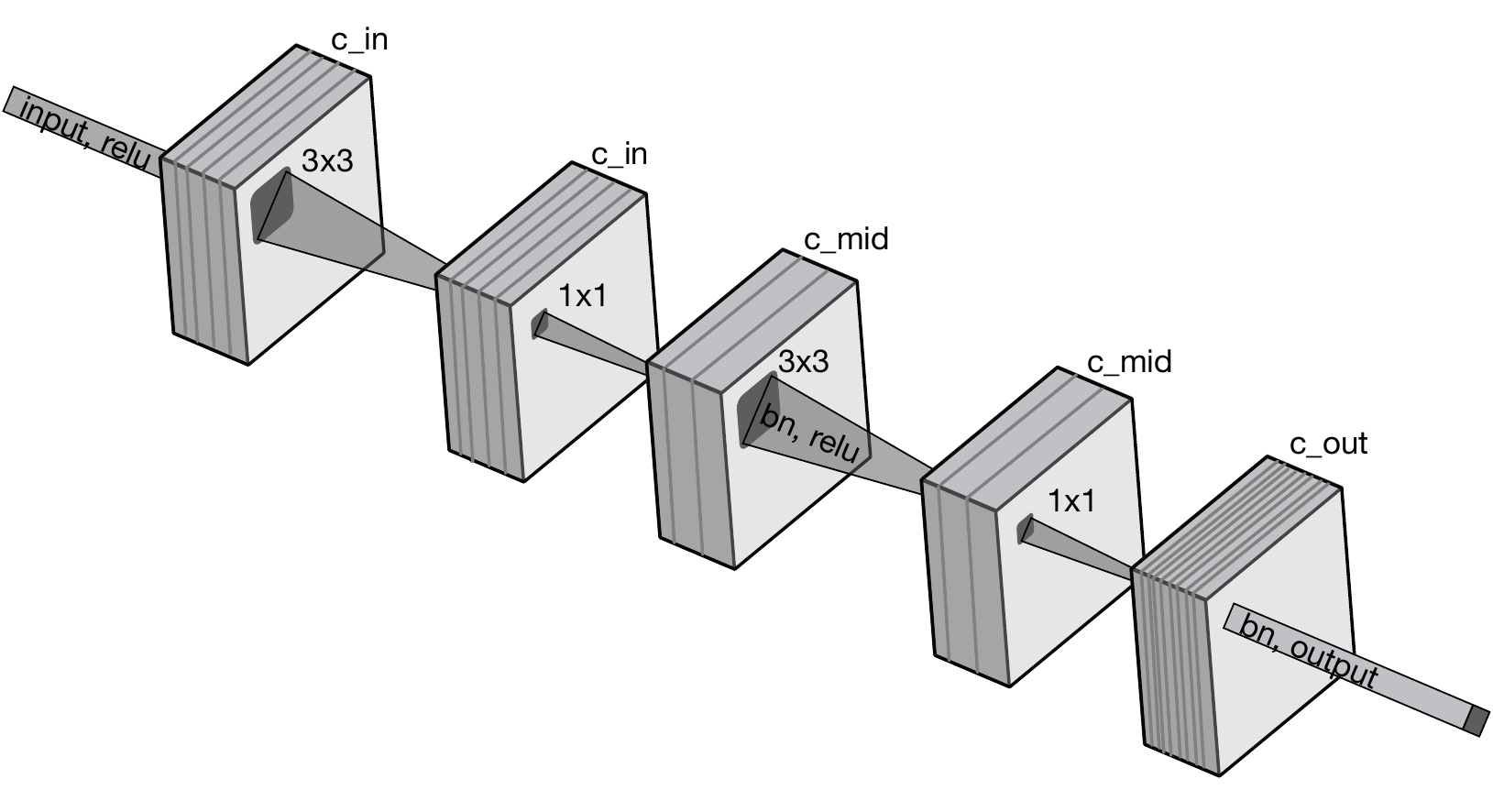}
    \caption{\small
    \label{fig:SharpSepConv} 
    Our \texttt{SharpSepConv} Block configured with 3x3 kernels and stride 1. All convolutions are either 1x1 convolutions or Depthwise Separable Convolutions\cite{chollet2017xception}. The \texttt{c\_mid} hyperparameter can be either proportional to \texttt{c\_out} or set to an arbitrary constant.
}
\vspace{-0.2cm}
\end{figure}

Neural Architecture Search (NAS) promises to automatically and efficiently optimize a complex model based on representative data so that it will generalize, and thus make accurate predictions on new examples.
Recent NAS results have been impressive, notably in computer vision, but the search process took large amounts of computing infrastructure~\cite{2017nasnet,zoph2017neural,liu2018PNAS}. 
These methods were followed up with a multiple order of magnitude increase in search efficiency~\cite{2018enas,liu2018darts}.
However, key questions remain:
What makes a search space worth exploring?
Is each model visited in a search space getting a fair shot?

We investigate how to improve the search space of DARTS\cite{liu2018darts}, which is one of several based on NASNet\cite{2017nasnet,real2018regularized,2018enas,liu2018PNAS,huang2018gpipe}, propose  Differentiable Hyperparameter Grid Search with DARTS, and draw conclusions that apply across architecture search spaces. 

\noindent
To summarize, we make the following contributions:
\begin{enumerate}
\item Define the novel \texttt{SharpSepConv} block with a more consistent structure of model operations and an adaptable middle filter count in addition to the sharpDARTS architecture search space. This leads to highly parameter-efficient results which match or beat state of the art performance for mobile-scale architectures on CIFAR-10, CIFAR-10.1, and ImageNet with respect to Accuracy, AddMult operations, and GPU search hours.
\item Introduce the Cosine Power Annealing learning rate schedule for tuning between Cosine Annealing\cite{loshchilov2016sgdrcosineannealing} and exponential decay, which maintains a more optimal learning rate throughout the training process.
\item Introduce Differentiable Hyperparameter Grid Search and the HyperCuboid search space for efficiently evaluating arbitrary discrete choices.
\item Demonstrate the low-capacity bias of DARTS search on two search spaces, and introduce Max-W Weight Regularization to correct the problem.
\end{enumerate}

\section{Related Work} 
\label{ref:related}

Architecture search is the problem of optimizing the structure of a neural network to more accurately solve another underlying problem. 
In essence, the design steps of a neural network architecture that might otherwise be done by an engineer or graduate student by hand are instead automated and optimized as part of a well defined search space of reasonable layers, connections, outputs, and hyperparameters. 
In fact, architecture search can itself be defined in terms of hyperparameters\cite{hundt2018hypertree} or as a graph search problem\cite{2017nasnet, 2018enas, cai2018proxylessnas, NIPS2016_ConvNeuralFabrics}. 
Furthermore, once a search space is defined various tools can be brought to bear on the problem including Bayesian optimization\cite{pmlr-v64-mendoza_towards_2016}, other neural networks\cite{smashHyperNetworks}, reinforcement learning, evolution\cite{real2017large, real2018regularized}, or a wide variety of optimization frameworks. 
A survey for the topic of Neural Architecture Search (NAS) is available at \cite{2018nassurvey}.

Differentiable Architecture Search (DARTS)\cite{liu2018darts} defines a search space in terms of parameters $\alpha$, weights $w_i = softmax(\alpha_i)$, and operation layers $op_i$  which are made differentiable in the form $output_i = w_i * op_i(input_i)$. 
ProxylessNAS\cite{cai2018proxylessnas} works in a similar manner on a MobileNetv2\cite{s2018mobilenetv2} based search space. 
It only loads two architectures at a time, updating $w$ based on relative changes between them. 
This saves GPU memory so that ImageNet size datasets and architectures load directly, but at the cost of shuttling whole architectures between the GPU and main memory.
\begin{figure}[btp!]
    \centering
    \includegraphics[width=1\columnwidth]{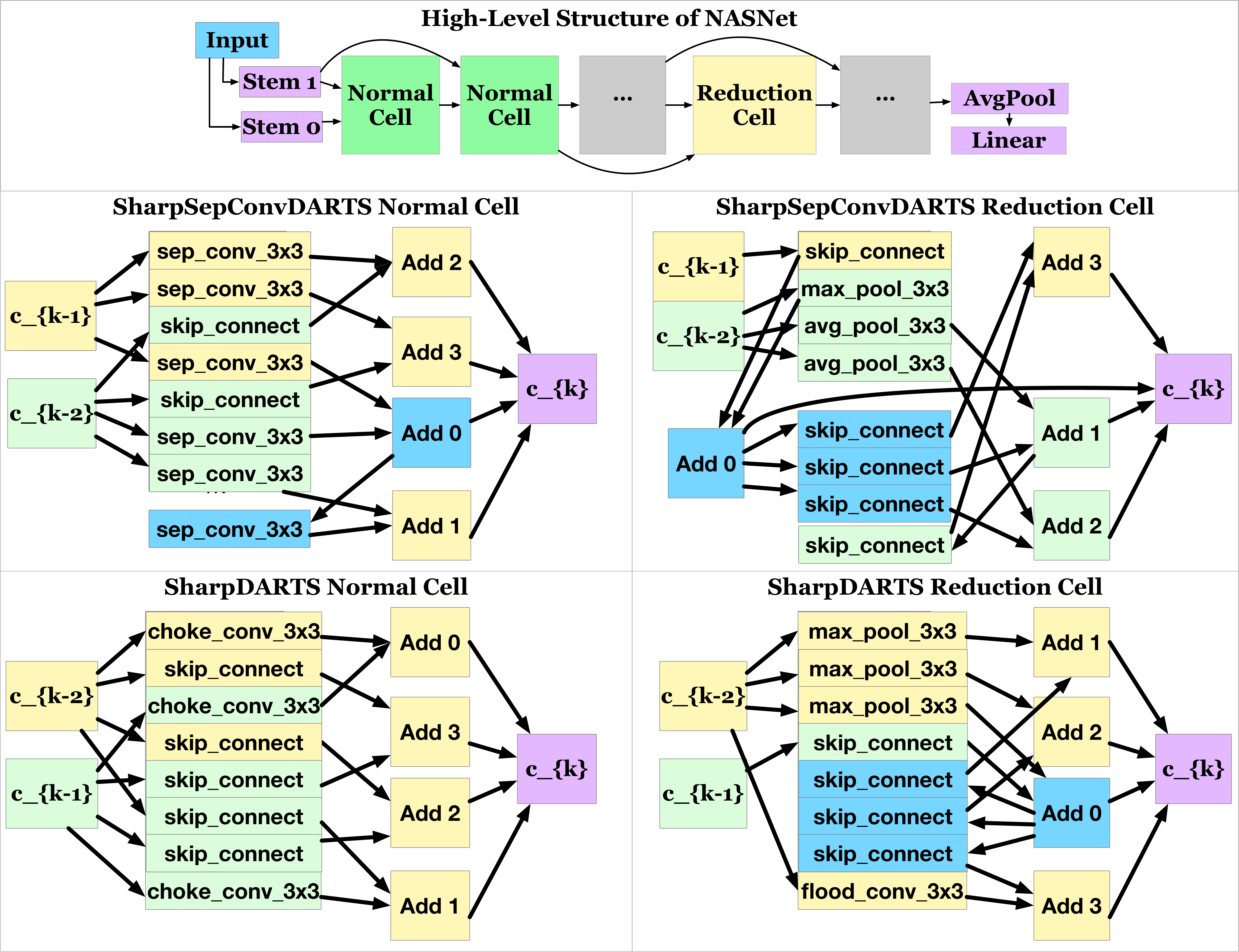}
    \caption{\small
    \label{fig:sharpdartscells} 
    (Top) High-level structure of NASNet\cite{2017nasnet} style architectures like sharpDARTS. Normal cells are stride 1 and Reduction cells are stride 2
    (Left) The SharpDARTS and SharpSepConvDARTS normal cell. (Right) The SharpDarts and SharpSepConvDARTS reduction cell.
}
\vspace{-0.2cm}
\end{figure}

Depthwise Separable Convolutions (SepConv) are a common building block of these searches, and were described as part of the Xception\cite{chollet2017xception} architecture, which improved efficiency on a per-parameter basis compared to its predecessor Inception-v3\cite{szegedy2016rethinking_incptionv3}. 
It was subsequently used to great effect in MobileNetV2\cite{s2018mobilenetv2}.
A SepConv is where an initial convolution is defined in which the number of groups is equal to the number of input channels, followed by a single 1x1 convolution with a group size of 1. 
This type of convolution tends to have roughly equivalent or better performance than a standard Conv layer with fewer operations and lower memory utilization.\cite{guo2018network, chollet2017xception}
Furthermore, so-called ``bottleneck'' layers or blocks have proven useful to limiting the size and improving the accuracy of neural network models\cite{he2015resnet, s2018mobilenetv2}.

Augmentation is another fundamental tactic when optimizing the efficiency of neural networks. 
For example, on CIFAR-10 Cutout\cite{2017cutout} randomly sets 16x16 squares in an input image to zero; and
AutoAugment\cite{cubuk2018autoaugment} is demonstrated on PyramidNet\cite{Han2017PyramidNet}, where it applies reinforcement learning to optimize parameter choices for a set of image transforms, and to the odds of applying each transform during training.

\begin{table}\centering
\setlength\tabcolsep{3pt}
\begin{tabular}{lllllr}
\toprule
\multicolumn{6}{c}{Operations in the sharpDARTS search space} \\
\cmidrule(r){1-6}
Name & K & S & D & CMid & CMidMult \\
\cmidrule(r){1-6}

none & -- & -- & -- & -- & --   \\
AvgPool3x3 & 1 & 1 & 1 & -- & -- \\
MaxPool3x3 & 1 & 1 & 1 & -- & -- \\
SkipConnect & 1 & 1 & 1 & -- & --  \\
SepConv3x3 & 3 & 1 & 1 & -- & 1   \\
DilConv3x3 & 3 & 1 & 2 & -- & 1    \\
FloodConv3x3 & 3 & 1 & 1 & -- & 4    \\
DilFloodConv3x3 & 3 & 1 & 2 & -- & 4    \\
ChokeConv3x3 & 3 & 1 & 1 & 32 & --    \\
DilChokeConv3x3 & 3 & 2 & 2 & 32 & --       \\
\bottomrule
\end{tabular}
\caption{\small
\label{table:searchspace}
Available operations for connections within cells in the sharpDARTS search space. 
Columns K, S, and D are kernel, stride, and dilation, respectively. All ``Conv'' operations are configurations of \texttt{SharpSepConv}, shown in Fig. \ref{fig:SharpSepConv} and \ref{fig:pytorch_sharpsepconv}.
}
\vspace{-0.2cm}
\end{table}
\begin{figure}[bt!]
%
%
\includegraphics[width=\columnwidth]{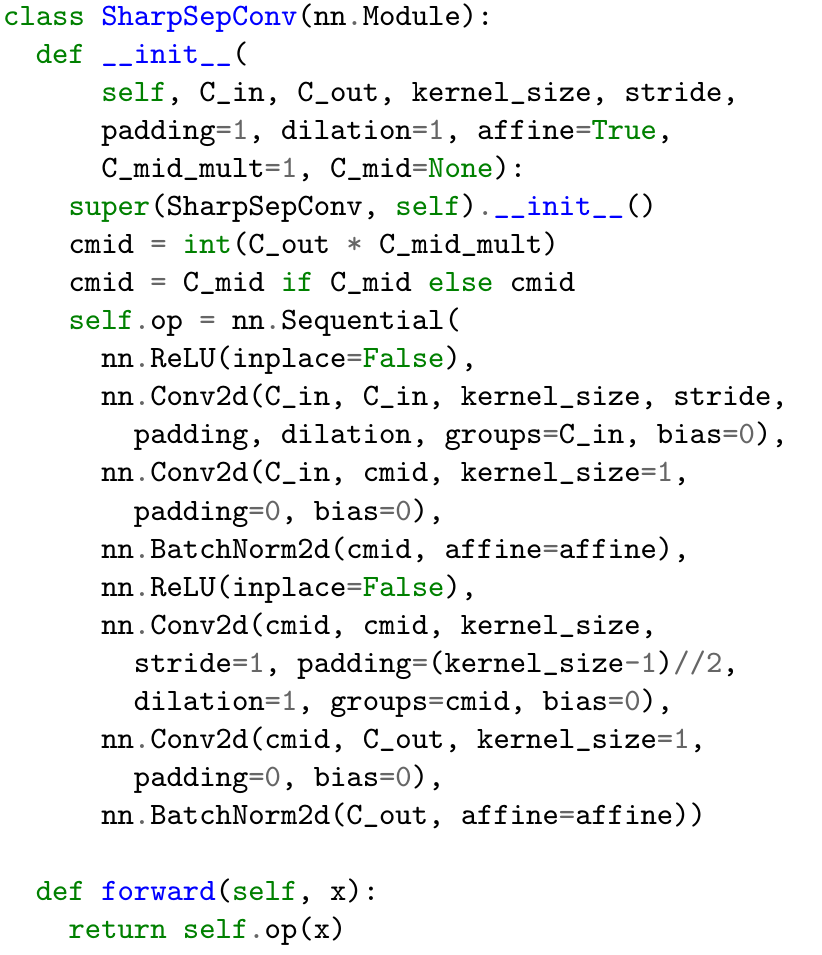}
\caption{
\small
\label{fig:pytorch_sharpsepconv}
The complete PyTorch 1.0~\cite{paszke2017pytorch} definition of the \texttt{SharpSepConv} block visualized in Fig. \ref{fig:SharpSepConv}.}
\vspace{-0.4cm}
\end{figure}


\begin{figure*}[bt!]
\vspace{-0.2cm}
    \centering
    \hfill
    \includegraphics[width=\textwidth]{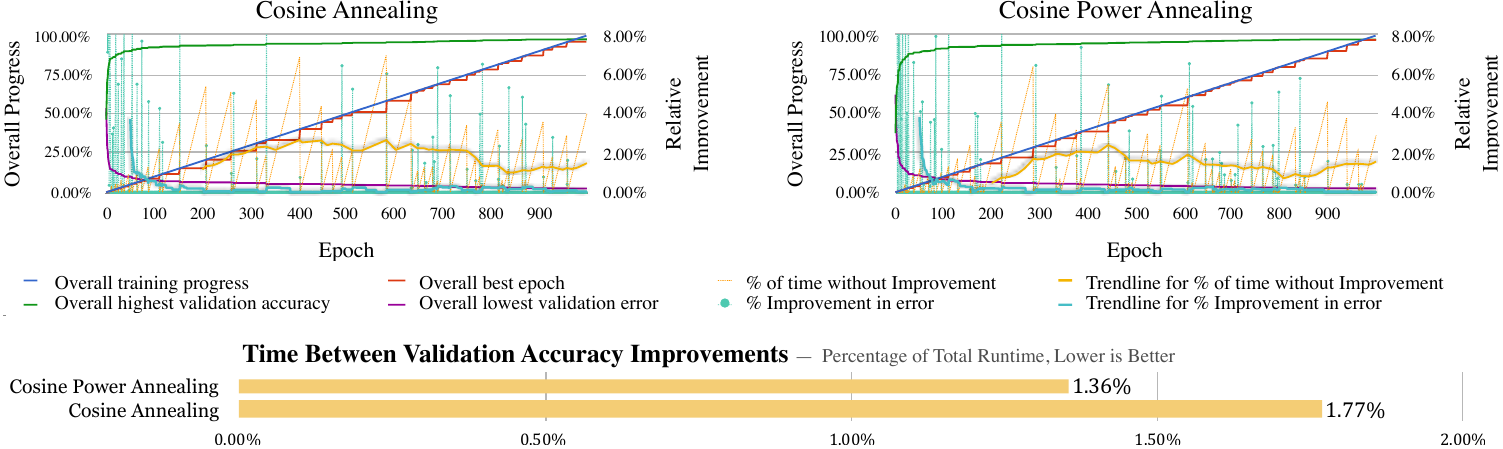}
    \caption{
    \label{fig:compare_cosine_annealing}
    \small
    A comparison of annealing methods. Cosine Power Annealing is more frequently at an optimal learning rate.  
    These 1000 epoch examples of CIFAR-10 sharpDARTS SGD training were selected to compare progress. Here final validation accuracy differs by 0.04\%, but in the typical case Cosine Power Annealing has better performance.
    }
\vspace{-0.4cm}
\end{figure*}
\begin{figure}[bt!]
    \centering
    \hfill
    \includegraphics[width=\columnwidth]{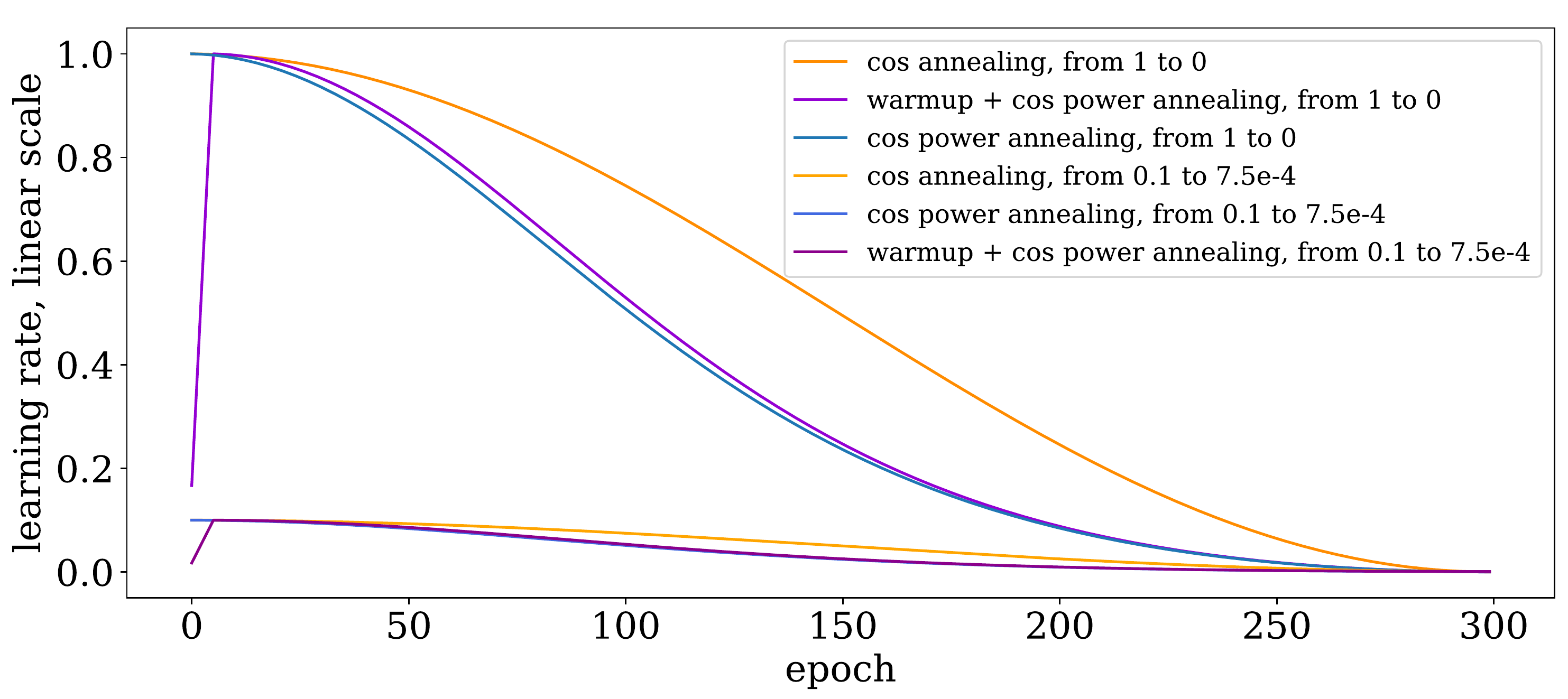}
    \includegraphics[width=\columnwidth]{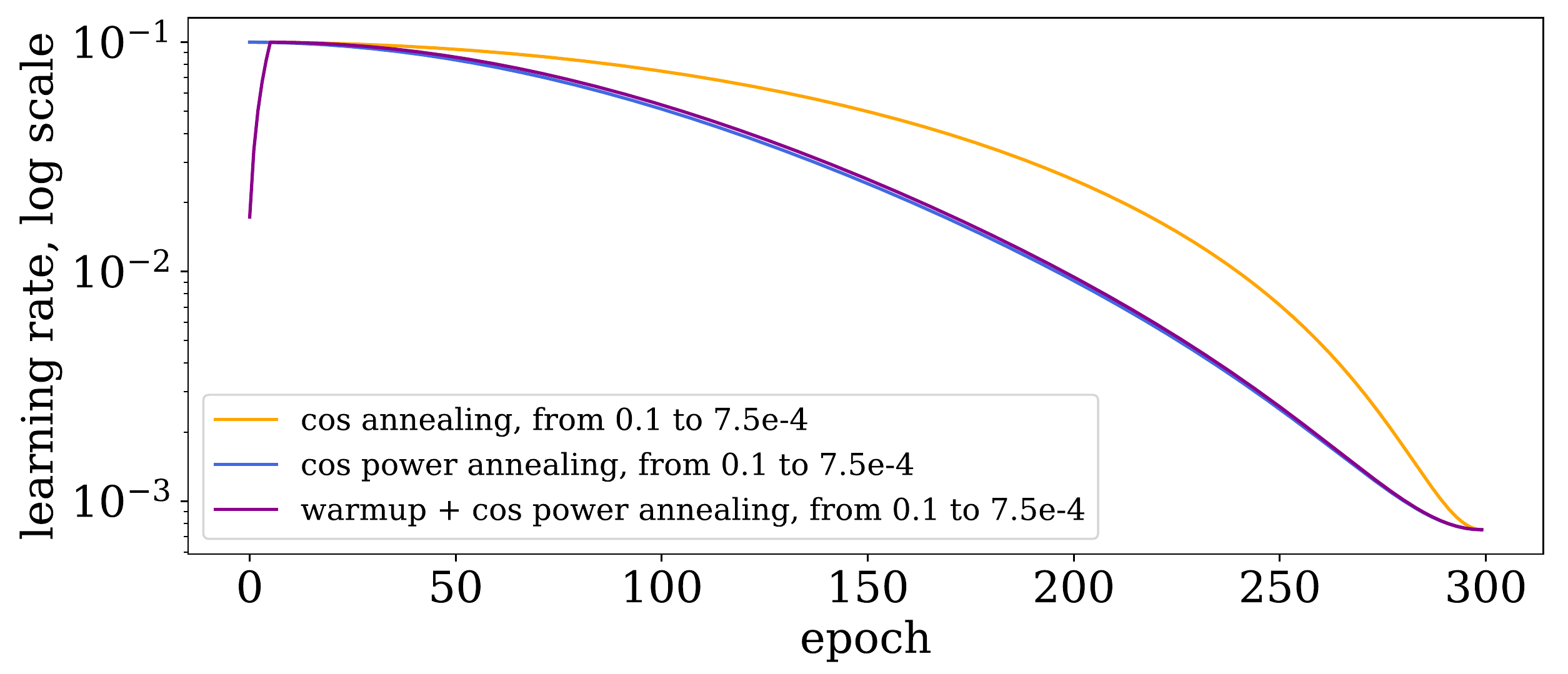}
    \caption{
    \label{fig:cosine_power_annealing}
    \small
    A comparison of Cosine Annealing~\cite{loshchilov2016sgdrcosineannealing} (Eq. \ref{eq:cosineannealing}) and Cosine Power Annealing (Eq. \ref{eq:cosinepowerannealing} with $p=10$) learning rate schedules for ImageNet. The bottom chart's logarithmic scale demonstrates how the cosine term is most prominent during early and late epochs, while the exponential term $p$ smooths decay through the middle epochs.
}
\vspace{-0.2cm}
\end{figure}

However, improvements via the methods above are only valuable if the results are reproducible and generalize to new data, issues which are a growing concern throughout academia. 
Recht et al.\cite{recht2018cifar10} have investigated such concerns, creating a new CIFAR-10.1 test set selected from the original tiny images dataset from which CIFAR-10 was itself selected. 
While the rank ordering of evaluated models remains very consistent, their work demonstrates a significant drop in accuracy on the test set relative to the validation set for all models.
Variation implicit in the training process, models, and the dataset itself must be carefully considered when reading results; and so a slightly better score does not guarantee better generalization.
For reference, on CIFAR-10.1 ``a conservative confidence interval (Clopper-Pearson at confidence level 95\%) for accuracy 90\% has size about $\pm1\%$ with n = 2,000 (to be precise, [88.6\%, 91.3\%])''\cite{recht2018cifar10}.
\section{Search Space and Training Methods}
We seek better generalization of architecture search using fewer resources. We begin by analyzing DARTS\cite{liu2018darts}, particularly its search space implementation and training regimen.
The search space covers a variety of core operations or ``blocks'' which include pooling, skip connections, no layer, separable convolutions and dilated convolutions.
Upon deeper analysis, we note that the \texttt{DilConv} operation contains 2 convolution layers while \texttt{SepConv} contains 4, thus comparing blocks with different scales.
This scale imbalance matters, and
Sec. \ref{sec:ablation_study}, \ref{sec:hypercuboids_and_regularization}, and \ref{sec:grid_search} will explore why, but first we will go through the design choices and data leading up to that conclusion.
Therefore, we start with our \texttt{SharpSepConv} block, which helps to correct the aforementioned imbalance, and the Cosine Power Annealing learning rate schedule.

\subsection{Sharp Separable Convolution Block}
\label{ssec:goalsandencodings}
We first define the \texttt{SharpSepConv} block which consists of 2 Separable Convolutions and 2 bottlenecks to balance the number of layers and to capitalize on these components' computational efficiency, as we described in Related Work (Sec. \ref{ref:related}).
All of the parameters of \texttt{SharpSepConv} are visualized in Fig. \ref{fig:SharpSepConv} and the block is defined using the PyTorch 1.0\cite{paszke2017pytorch} framework in Fig. \ref{fig:pytorch_sharpsepconv}. 
We will refer to these figures and the definitions in the Table \ref{table:cifar10} caption throughout the remaining text.

The \texttt{SharpSepConv} block permits variation in the dilation parameters and the number of filters contained in the middle layers, both relative to the input and in absolute terms, without changing the number of convolutions. 
In effect, \texttt{C\_mid} adds fixed-size bottlenecks directly into the architecture search space which can have a large impact on the number of AddMult operations in an architecture and thus the computational efficiency. 
Furthermore, \texttt{C\_mid\_mult} makes it possible to incorporate an additional reduction or increase in network size as needed within individual cells as can be seen in Fig. \ref{fig:sharpdartscells}.

Using \texttt{SharpSepConv} and other operations we create the sharpDARTS search space defined in Table \ref{table:searchspace}.
As the ablation study in Sec. \ref{sec:ablation_study} and the figures in Table \ref{table:cifar10} show, \texttt{SharpSepConv} operations substantially contribute to the improvements inherent to the final sharpDARTS and SharpSepConvDARTS architectures.
%
%

\begin{table*}\centering
\ra{1.3}
\setlength\tabcolsep{3pt}
\begin{tabular}{@{}lccccccccr@{}}\toprule
\multicolumn{1}{l}{Architecture} & \multicolumn{1}{c}{Auto}   & \multicolumn{1}{c}{Grad}   & \multicolumn{1}{c}{SSC}   & \multicolumn{1}{c}{Val Error}  & \multicolumn{1}{c}{Test Error}  & \multicolumn{1}{c}{Par.} & \multicolumn{1}{c}{$+ \times$} & \multicolumn{1}{c}{GPU} &
\multicolumn{1}{c}{Algorithm}\\
\multicolumn{1}{l}{}   & \multicolumn{1}{c}{Aug}   & \multicolumn{1}{c}{Order}   & \multicolumn{1}{c}{}   & \multicolumn{1}{c}{CIFAR-10}  & \multicolumn{1}{c}{CIFAR-10.1}  & \multicolumn{1}{c}{M} & \multicolumn{1}{c}{B} & \multicolumn{1}{c}{Days} &
\multicolumn{1}{c}{}\\
\midrule
ShakeShake64d\cite{gastaldi2017shakeshake, recht2018cifar10}&\xmark&&\xmark&2.9$\pm$0.3 &7.0$\pm$1.8&26.2&&--&Man\\
AutoAugment\cite{cubuk2018autoaugment,Han2017PyramidNet}&\checkmark&&\xmark& 1.48 && 26.0 && -- & AugRL+Man \\
GPipe AmoebaNet-B\cite{huang2018gpipe,real2018regularized}&\xmark&&\xmark& 1.0 $\pm$ 0.05 && $\sim$557 &$\sim$50&& Evo+Man \\\midrule
NASNet-A \cite{2017nasnet}  &\xmark&&\xmark& 2.65    && 3.3 && 1800 & RL \\
NASNet-A  \cite{2017nasnet}$^\dagger$ &\xmark&&\xmark& 2.83    && 3.8 &0.624& 3150 & RL \\ 
AmoebaNet-A \cite{real2018regularized}  &\xmark&&\xmark& 3.34 $\pm$ 0.06    && 3.2 && 3150 & Evo \\ 
AmoebaNet-A \cite{real2018regularized}$^\dagger$&\xmark&&\xmark& 3.12    && 3.1 &0.506& 3150 & Evo \\
AmoebaNet-B \cite{real2018regularized}&\xmark&&\xmark& 2.55 $\pm$ 0.05 && 2.8 && 3150 & Evo \\
ProgressiveNAS \cite{liu2018PNAS}&\xmark&&\xmark& 3.41 $\pm$ 0.09 && 3.2 && 225 & SMBO \\
ENAS \cite{2018enas}&\xmark&&\xmark& 2.89    && 4.6 && 0.5 & RL \\
DARTS random~\cite{liu2018darts}&\xmark&&\xmark& 3.49                         && 3.1      &&    --     & --       \\ 
DARTS$^\dagger$~\cite{liu2018darts}&\xmark&1&\xmark& 2.94 && 2.9 &0.518& 1.5 &Grad\\
DARTS$^\dagger$~\cite{liu2018darts}&\xmark&2&\xmark& 2.83 $\pm$ 0.06 &&  3.4  &0.547& 4 &Grad\\
ProxylessNAS~\cite{cai2018proxylessnas}&\xmark&1&\xmark& 2.08 && 5.7 && 4 & Grad\\
\midrule
\textbf{SharpSepConvDARTS}&\checkmark&\textbf{1}&\checkmark& \textbf{1.98$\pm$0.07}  &\textbf{5.9$\pm$0.5}& \textbf{3.6} &\textbf{0.579}&\textbf{0.8}&\textbf{Grad}\\
\textbf{DARTS+SSC, no search}&\checkmark&\textbf{2}&\checkmark& \textbf{2.05$\pm$0.14} &\textbf{5.6$\pm$0.8}& \textbf{3.5} &\textbf{0.576}& -- &\textbf{Grad}\\
\textbf{sharpDARTS}&\checkmark&\textbf{1}&\checkmark& \textbf{2.29$\pm$0.06} &\textbf{6.2$\pm$0.4}& \textbf{1.98} &\textbf{0.357}& \textbf{1.8} &\textbf{Grad}  \\
SharpSepConvDARTS&\xmark&1&\checkmark& 2.45  && 3.6 &0.579& 0.8 &Grad\\
DARTS+SSC, no search&\xmark&2&\checkmark& 2.55 && 3.5 &0.576& -- &Grad\\
sharpDARTS&\xmark&1&\checkmark& 2.97  && 1.98 &0.357& 1.8 &Grad\\ \bottomrule
\bottomrule
\end{tabular}
\caption{\small
\label{table:cifar10} 
Results for the CIFAR-10 dataset and the CIFAR-10.1\cite{recht2018cifar10} test set, \textbf{bold lines} are trained with AutoAugment\cite{cubuk2018autoaugment} and Cosine Power Annealing (Eq. \ref{eq:cosinepowerannealing}) where we set $p=2$, $\eta^0_{max}=0.025$, $\eta^0_{min}=1\text{e-}8$, and $T_0=2000$ epochs of training for our best results.
Our figures with a range in the bottom section are defined as $median \pm range/2$ and include at least 3 runs with Cosine Power Annealing.
Dagger$^\dagger$ indicates genotypes available in our accompanying repository. 
Non-bold numbers have 1000 epochs of Cosine Power Annealing and AutoAugment disabled to better match figures from past work.
The code and this table is derived from DARTS\cite{liu2018darts}.
The \textbf{GradOrder} column is 1 for standard gradients, 2 for Hessians on DARTS based algorithms.
All algorithms utilize \textbf{CutOut}\cite{2017cutout} except for ProgressiveNAS.
A \textbf{``genotype''} simply defines an architecture as a graph. It has a list of layers to use, and the previous node, which it should use as an input. A \textbf{``primitive''} is an operation aka block of layers that can be chosen during the search. A \textbf{``node''} takes the output of two instances of an (input, primitive) pair as input and adds them, as seen in Fig. \ref{fig:sharpdartscells}. 
\textbf{DARTS+SSC, no search} compares the genotype published by DARTS with \texttt{SharpSepConv} layers in the primitives. \textbf{SharpSepConvDARTS} is the same as DARTS + SharpSepConv but uses a genotype determined by a new search incorporating the DARTS primitives with SharpSepConv blocks. 
sharpDARTS uses the architecture search space described in Table \ref{table:searchspace}.}
\vspace{-0.2cm}
\end{table*}
\begin{table*}\centering
\ra{1.3}
\setlength\tabcolsep{2pt}
	\centering
	\begin{tabular}{@{}lccccccc@{}}
		\toprule
 		\multirow{2}{*}{\textbf{Architecture}} & \multicolumn{2}{c}{\textbf{\% Test Error}}      & \textbf{Params} & \textbf{$+ \times$}  & \textbf{Search Cost}  & \textbf{Search} \\ \cline{2-3}
 		& top-1 & top-5      & \textbf{M} & \textbf{B}  & \textbf{GPU Days} & \textbf{Method} \\ \midrule
		GPipe AmoebaNet-B \cite{real2018regularized,huang2018gpipe} & 15.7 & 3.0 & 557 & $\sim$50 & 3150 & Evo+Man \\ 
		MobileNetv2 \cite{s2018mobilenetv2} & 28.0 & -- & 3.4 & 0.300 & -- & Man \\ 
		MobileNetv2 (1.4) \cite{s2018mobilenetv2} & 25.3 & -- & 6.9 & 0.585 & -- & Man \\ 
    \midrule
		NASNet-A \cite{2017nasnet} & 26.0 & 8.4 & 5.3 & 0.564 & 1800 & RL \\
		NASNet-B \cite{2017nasnet} & 27.2 & 8.7 & 5.3 & 0.488 & 1800 & RL \\
		AmoebaNet-B \cite{real2018regularized} & 26.0 & 8.5 & 5.3 & 0.555 & 3150 & Evolution \\
		AmoebaNet-C \cite{real2018regularized} & 24.3 & 7.6 & 6.4 & 0.570 & 3150 & Evolution \\
		ProgressiveNAS \cite{liu2018PNAS} & 25.8 & 8.1 & 5.1 & 0.588 & $\sim$225 & SMBO \\
		DARTS \cite{liu2018darts}    &  26.9   &  9.0 & 4.9 & 0.595 & 4 & Grad \\ 
		ProxylessNAS \cite{cai2018proxylessnas} & 24.9 & 8.1 & 5.1 & 0.581 & $\sim$8 & Grad \\\midrule
		\textbf{SharpSepConvDARTS} &  \textbf{25.1}  & \textbf{7.8} & \textbf{4.9} & \textbf{0.573} & \textbf{0.8} & \textbf{Grad} \\
		\textbf{sharpDARTS cmid 96}    &  \textbf{27.8}  &  \textbf{9.2} & \textbf{3.71} & \textbf{0.481} & \textbf{1.8} & \textbf{Grad} \\
		\textbf{sharpDARTS cmid 32}    &  \textbf{31.3}  &  \textbf{11.34} & \textbf{3.17} & \textbf{0.370} & \textbf{1.8} & \textbf{Grad} \\

		\bottomrule
	\end{tabular}
	\caption{\small
	Mobile ImageNet Architecture Comparison, table updated from DARTS\cite{liu2018darts}. Lower values are better in all columns. A dash indicates data was either not available or not applicable.
	\label{table:imagenet}}
\vspace{-0.2cm}
\end{table*}

\subsection{Cosine Power Annealing}

Cosine Annealing\cite{loshchilov2016sgdrcosineannealing} is a method of adjusting the learning rate over time, reproduced below:
\begin{eqnarray}
	\label{eq:cosineannealing}
	\eta_t = \eta^i_{min} + \frac{1}{2}(\eta^i_{max} - \eta^i_{min}){
	\left(1 + \cos\left(\pi \frac{T_{cur}}{T_i}\right)\right)}	
\end{eqnarray}
Where $\eta^i_{min}$ and $\eta^i_{max}$ are the minimum and maximum learning rates, respectively; $T_i$ is the total number of epochs; $T_{cur}$ is the current epoch; and $i$ is the index into a list of these parameters for a sequence of warm restarts in which $\eta^i_{max}$ typically decays.
This schedule has been widely adopted and it is directly implemented in PyTorch 1.0~\cite{paszke2017pytorch} without warm restarts, where $i = 0$.

The Cosine Annealing schedule works very well and is employed by DARTS.
However, as can be seen in Fig. \ref{fig:compare_cosine_annealing}, the average time between improvements is above 2\% of the total runtime between epochs 300 and 700.
This is an imbalance in the learning rate, which is an artifact of the initial slow decay rate of Cosine Annealing followed by the rapid relative decay in learning rate. 
We mitigate this imbalance by introducing a power curve parameter $p$ into the algorithm which we call \textbf{Cosine Power Annealing:}
\begin{equation}
	\label{eq:cosinepowerannealing}
	\eta_t = \eta^i_{min} + (\eta^i_{max} - \eta^i_{min})\frac{p^{\frac{1}{2}
	\left(1 + \cos\left(\pi \frac{T_{cur}}{T_i}\right)\right) + 1}-p}{p^2-p}
\end{equation}

The introduction of the normalized exponential term $p$ permits tuning of the curve's decay rate such that it maintains a high learning rate for the first few epochs, while simultaneously taking a shallower slope during the final third of epochs.
These two elements help reduce the time between epochs in which validation accuracy improves. A comparison is provided in Fig. \ref{fig:compare_cosine_annealing} and \ref{fig:cosine_power_annealing}. 
In our implementation we also define a special case for the choice of $p=1$ such that it falls back to standard cosine annealing.
This algorithm is also compatible with decaying warm restarts, but we leave that schedule out of scope for the purposes of this paper.


\subsection{Results}

\label{sec:results}

\textbf{CIFAR-10 and CIFAR-10.1:} 
Our absolute top performance is \textbf{SharpSepConvDARTS} with 1.93\% CIFAR-10 top-1 validation error (1.98$\pm$0.07) and 5.925$\pm$0.48 CIFAR-10.1 test error (Table \ref{table:cifar10}) when including our improved training regimen. 
To the best of our knowledge, this is state of the art performance for mobile-scale ($\sim$600M AddMult ops\cite{liu2018darts}) and ProxylessNAS\cite{cai2018proxylessnas} is in second at 2.08\% val error. 
We also show a statistically significant\cite{recht2018cifar10} improvement over ShakeShake64d\cite{recht2018cifar10,gastaldi2017shakeshake} which is the best available CIFAR-10.1 model, with 7.0$\pm$1.2 test error.

GPipe AmoebaNet-B\cite{huang2018gpipe} remains the best at any scale with 1\% val error. 
It is scaled up from the original AmoebaNet-B and we expect other models to scale in a similar way.
This truly massive model cannot load on typical GPUs due to 557M parameters and billions of AddMult ops. 
It runs with specialized software across multiple Google TPU hardware devices. 

Our training time\footnote{This research was conducted on 5 Nvidia GPU types: Titan X, GTX 1080 Ti, GTX 1080, Titan XP, and the RTX 2080Ti.} for a 60 epoch search is 0.8-1.2 days, with 2k epochs of mixed fp16 training of a final model in 1.7-2.8 days, totaling 2.9-3.6 GPU-days end-to-end on one RTX 2080Ti.
The discrepancy in totals is because the slower sharpDARTS search finds a faster final model.
\textbf{ImageNet:}
Our top \textbf{SharpSepConvDARTS} model achieved 25.1\% top-1 and 7.8\% top-5 error, which is competitive with other state of the art mobile-scale models depicted in Table \ref{table:imagenet}, and this translates to relative improvements over DARTS of 7\% in top-1 error, 13\% in top-5 error, and 80\% in search time.
Our ImageNet model uses the genotype of the CIFAR-10 search and follows the same cell based architecture of \cite{2017nasnet, liu2018darts, 2018enas} with different operations in our search space.
We apply random cropping to 224x224, random horizontal flipping, AutoAugment\cite{cubuk2018autoaugment}, normalization to the dataset mean and std deviation, and finally Cutout\cite{2017cutout} with a cut length of 112x112.
Training of final models was done on 2x GTX 2080Ti in 16 bit mixed precision mode and takes 4-6 days, which is 8-12 GPU-days, depending on the model.

\section{Towards Better Generalization}
\label{sec:towards_better_generalization}
DARTS search improved results over random search by 19\% (Table \ref{table:cifar10}), and by adding our training regimen and search space improvements we get an additional 30\% relative improvement over DARTS.
Manual changes like those made to the \textbf{SharpSepConv} block and to AmoebaNet-B for GPipe\cite{huang2018gpipe} are not represented in any search space, and yet they directly lead to clear improvements in accuracy.
So why aren't they accounted for?
Let's assume that it is possible to encode all of these elements and more into a single, broader search space in which virtually every neural network graph imaginable is encoded by hyperparameters.
To even imagine tackling a problem of this magnitude, we must first ask ourselves an important question:
Does DARTS even generalize to other search domains designed with this challenge in mind?
In this section, we provide a preliminary exploration to begin answering these questions.

\begin{figure*}[btp!]
    \centering
    \hfill
    \includegraphics[width=0.95\textwidth]{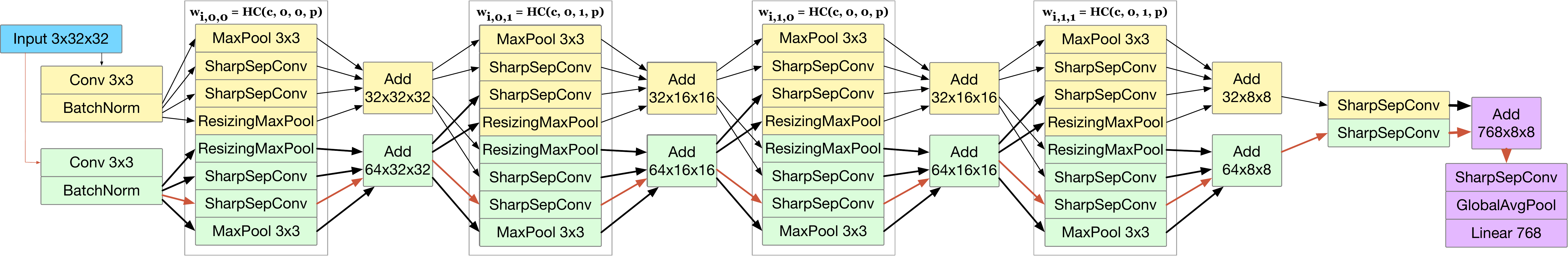}
    \caption{\small
    \label{fig:multichannel} 
    A HyperCuboid with shape HC = (filter\_scales, normal\_layers, reduction\_layers, layer\_types) = ((2x2)x2x2x2) for Differentiable Hyperparameter Grid Search (Sec. \ref{sec:grid_search}).
    The data is ordered $(channels, width, height)$.
    An example single-path model is highlighted with red arrows. 
    Here you can see a small scale n-tuple coordinate system which might serve as the conceptual basis for future large scale search spaces. (Sec. \ref{sec:towards_better_generalization})}
\vspace{-0.6cm}
\end{figure*}

\subsection{Differentiable Grid Search}
\label{sec:grid_search}
We introduce \textbf{Differentiable Hyperparameter Grid Search} which is run on a \textbf{HyperCuboid Search Space} parameterized by an n-tuple, such as $HC = (\gamma,d_n,d_r,p)$, where $\gamma$ is an arbitrary set of hyperparameters; $d_n$ is the number layers in one block; $d_r$ is the number of layer strides; and $p$ is an arbitrary set of choices, primitives in this case.
In the HyperCuboid Search Space many possible paths pass through each node, and all final sequential paths in the Directed Acyclic Graph of architecture weights are of equal length.
The number of architecture weights $w$ in the HyperCuboid graph is the product of the size of the tuple elements; and the number of hyperparameters, primitives, and tuple size can vary in this design.
We test a specific HyperCuboid called \textbf{MultiChannelNet} with a tuple (filter\_scale\_pairs, normal\_layer\_depth, reduction\_layer\_depth, primitives), visualized at a small ((2x2)x2x2x2) scale in Fig. \ref{fig:multichannel} with SharpSepConv and MaxPool primitives. 
Here $\gamma=2^n\times2^m|n,m\in[5,6]$ are combinations of possible input and output filter scales.
Our actual search has dimension ((4x4)x3x3x2)$|n,m\in[5..8]$ where final linear paths have 14 nodes, since ``Add'' nodes are excluded.
In MultiChannelNet, graph nodes determine which filter input, filter output, and layer type should be utilized.
Larger weights $w$ imply a better choice of graph node and we therefore search for an optimal path through a sequence of primitive weights which maximizes the total path score.

We construct a basic handmade model in one shot with \texttt{SharpSepConv} and 32 filters, doubling the number of output filters at layers of stride 2 up to the limit of 256. 
We also ran an automated DARTS search for 60 epochs and about 16 GPU-Hours to find an optimal model.
To our surprise, the handmade model outperforms DARTS by over 2\%, as Table \ref{table:multichannelnet_search_algorithms} indicates.
Why might this be? To answer this we return to our original NASNet based search space to look for discrepancies in an ablation study.

\begin{table}\centering
\vspace{0.2cm}
\setlength\tabcolsep{2pt}
\begin{tabular}{lrrcc}
\toprule
Weights/Path & Par. & $+ \times$ & Val Err \% & Test Err \% \\
\cmidrule(r){1-5}
Scalar&0.8M& 23M& 8.62$\pm$0.16& 17.2$\pm$0.8 \\ 
Handmade& 1.0M&25M& 6.50$\pm$0.18& 13.9$\pm$0.4 \\
\textbf{Max-W}&\textbf{0.9M}&\textbf{31M}&\textbf{5.44$\pm$0.12}&\textbf{12.4$\pm$1.0}\\ 
\bottomrule
\end{tabular}
\caption{\small
\label{table:multichannelnet_search_algorithms}
CIFAR-10 and CIFAR-10.1 results for model paths from our MultiChannelNet (Fig. \ref{fig:multichannel}) example of the HyperCuboid Search Space. 
There is a statistically significant\cite{recht2018cifar10} improvement from the original Scalar DARTS to our proposed Max-W Weight Regularization DARTS. 
We also see the hypothesized increase in accuracy, parameters, and AddMult ($+ \times$) flops described in Sec. \ref{sec:hypercuboids_and_regularization} and \ref{sec:grid_search}.
Final training is $\sim$0.3 days on an RTX 2080Ti.
}
\vspace{-0.3cm}
\end{table}

\subsection{Ablation Study}
\label{sec:ablation_study}
Figures for our ablation study are in Table \ref{table:cifar10} which also has a description of our preprocessing changes in the caption.
We will analyze the 3 main model configurations below:
\textbf{(1)} \textbf{DARTS+SSC} directly replaces all convolution primitives in DARTS\cite{liu2018darts} with a \texttt{SharpSepConv} layer where the block parameters, primitives, and the genotype are otherwise held constant; we see a 10\% relative improvement over DARTS val err (2.55\% vs 2.83\%) with only 5\% more AddMult operations and no additional augmentation.
\textbf{(2)} \textbf{SharpSepConvDARTS} is the same as item 1 but with a 1st order gradient DARTS search; we see relative improvements of 13\% in val err (2.45\% vs 2.83\%) and 80\% in search time.
\textbf{(3)} \textbf{sharpDARTS} (Table \ref{table:searchspace}) has slightly different primitives including \textit{flood} where middle channels expand 4x, \textit{choke} with a fixed 32 middle filters, and one of each conv with a \textit{dilation} of 2.

Our sharpDARTS model (Fig. \ref{fig:sharpdartscells}) with our improved training regimen achieved an absolute best error of 2.27\%.
Without training enhancements the 1st order gradient search of sharpDARTS has similar accuracy to 1st order DARTS and is definitively more efficient with 32\% fewer parameters, 31\% fewer AddMult operations, and lower memory requirements.
However, the absolute accuracy of sharpDARTS is \textit{marginally lower} than the original DARTS, and also suffers from a larger disparity on ImageNet (Table \ref{table:imagenet}). 
This is startling for two reasons:
(1) Substituting the \texttt{SharpSepConv} block improves accuracy in DARTS+SSC and SharpSepConvDARTS.
(2) The sharpDARTS search space still contains all primitives needed to represent both the final DARTS+SSC and SharpSepConvDARTS model genotypes perfectly.

We've replicated a discrepancy in DARTS behavior across two different search spaces, so the most likely remaining possibility must be a limitation in the DARTS search method itself.

\subsection{Max-W Regularization}
\label{sec:hypercuboids_and_regularization}

GPipe\cite{huang2018gpipe} shows AmoebaNet-B models improving in accuracy as they scale to $>$500M parameters and billions of AddMult flops (Fig. \ref{table:cifar10}, \ref{table:imagenet}).
These models are from a search space similar to those used by DARTS, among others\cite{liu2018PNAS,2018enas,2017nasnet}.
Assume that the GPipe scaling principle holds for similar training configurations and search spaces, and one might expect that DARTS would tend towards larger models throughout the search process.
However, during the early epochs of training, DARTS reliably produces models composed entirely of max pools and skip connects. 
These are among the smallest primitives in the DARTS architecture search space with respect to parameters and AddMult operations. 
Higher capacity layers are chosen later in the search process, as visualized in the animation included with the original DARTS source 
code\footnote{\scriptsize DARTS\cite{liu2018darts} model search animation for reduce cells: \url{https://git.io/fjfTC}  normal cells: \url{https://git.io/fjfbT}}.
We experimented with removing max pool layers and the undesired behavior simply shifts to skip connects.

We posit that the \textbf{DARTS Scalar Weighting} $output_i = w_i * op_i(input_i)$ tends towards the layers with the maximum gradient, and thus models will consist of smaller layers than appropriate.
Such bias during early phases of training is inefficient with respect to optimal accuracy, even if models might eventually converge to larger, more accurate models after a long period of training.
Therefore, we hypothesize that subtracting the maximum weight in a given layer will regularize weight changes via \textbf{Max-W Weighting:}
\begin{equation}
output_i = (1 - max(w) + w_i) * op_i(input_i)
\label{eqn:max_w}
\end{equation}
Intuitively, consider the architecture parameters $\alpha$, weights $w=softmax(\alpha) | w\in[0,1]$, highest score index $i_{max}=argmax(w)$,
and the highest score weight $w_{max}=w[i_{max}]$ at some arbitrary time during training.
If we apply Max-W weighting to the layer corresponding with $w_{max}$, we will have $(1 - w_{max} + w_{max}) * op_i(input_i)$ which reduces to $output_i = 1 * op_i(input_i)|i=i_{max}$. 
In this case the $\alpha$ underlying $w_{max}$ will remain unchanged, but this is not true for other values $w_i| i\neq i_{max}$.
Here it will be incumbent on non-maximum layers $w_i$ to outperform the highest score weight and grow their value $\alpha_i$. 
As values other than $w_{max}$ grow, $w_{max}$ will naturally drop in accordance with the behavior of $softmax$.
This has the net effect of reducing bias corresponding to the highest score layer $op_i$ paired with $w[i_{max}]$.

Our search with Max-W weighting on MultiChannelNet found a model which is both larger and significantly more accurate than both the original DARTS Scalar search models and a hand designed model (Table \ref{table:multichannelnet_search_algorithms}).
These results indicate that our initial hypothesis holds and Max-W DARTS (Eq. \ref{eqn:max_w}) is an effective approach to regularization when compared to standard Scalar DARTS.
Specific models will be released with the code.

\section{Future Work}

Our investigation also indicates several other areas for future work.
The SharpSepConvDARTS and sharpDARTS search spaces might also benefit from Max-W regularization, so it is an interesting topic for an additional ablation study.
MultiChannelNet and sharpDARTS indicate that the final DARTS model\cite{liu2018darts} did not fully converge to the optimum due to the scalar weighting bias (Sec. \ref{sec:ablation_study}, \ref{sec:hypercuboids_and_regularization}). 
We suspect other DARTS algorithms, such as ProxylessNAS\cite{cai2018proxylessnas}, suffer from this same bias.
We have also shown how Max-W Regularization correctly chooses larger models when those are more accurate, however this means a search with Max-W DARTS currently exceeds mobile-scale on the DARTS and sharpDARTS search spaces.
Adding a resource cost based on time and memory to each node might make it possible to directly optimize cost-accuracy tradeoffs with respect to a specific budget.
Arbitrary multi-path subgraphs respecting this budget could be chosen by iterative search with a graph algorithm like network simplex\cite{orlin1997polynomialnetworksimplex}.
Other alternatives include a reinforcement learning algorithm or differentiable metrics like the latency loss in ProxylessNAS\cite{cai2018proxylessnas}.


\section{Conclusion}

In this paper we met or exceeded state of the art mobile-scale architecture search performance on CIFAR-10, CIFAR-10.1 and ImageNet with a new \texttt{SharpSepConv} block.
We introduced the Cosine Power Annealing learning rate schedule, which is more often at an optimal learning rate than Cosine Annealing alone, and demonstrated an improved sharpDARTS training regimen.
Finally, we introduced Differentiable Hyperparameter Grid Search with a HyperCuboid search space to reproduce bias within the DARTS search method, and demonstrated how Max-W regularization of DARTS corrects that imbalance.

Finally, Differentiable Hyperparameter Search and HyperCuboids might be evaluated more broadly in the computer vision space and on other topics such as recurrent networks, reinforcement learning, natural language processing, and robotics.
For example, \texttt{SharpSepConv} is manually designed so a new HyperCuboid might be constructed to empirically optimize the number, sequence, type, layers, activations, normalization, and connections within a block.
Perhaps a future distributed large scale model might run Differentiable Hyperparameter Search over hundreds of hyperparameters which embed a superset of search spaces, making it possible to efficiently and automatically find new models for deployment to any desired application.

{\small
\bibliographystyle{style/ieee}
\bibliography{sample}

\begin{thebibliography}{10}\itemsep=-1pt

\bibitem{smashHyperNetworks}
A.~Brock, T.~Lim, J.~M. Ritchie, and N.~Weston.
\newblock {SMASH:} one-shot model architecture search through hypernetworks.
\newblock {\em CoRR}, abs/1708.05344, 2017.

\bibitem{cai2018proxylessnas}
H.~Cai, L.~Zhu, and S.~Han.
\newblock Proxyless{NAS}: Direct neural architecture search on target task and
  hardware.
\newblock In {\em International Conference on Learning Representations}, 2019.

\bibitem{chollet2017xception}
F.~Chollet.
\newblock Xception: Deep learning with depthwise separable convolutions.
\newblock {\em The IEEE Conference on Computer Vision and Pattern Recognition
  (CVPR)}, Jul 2017.

\bibitem{cubuk2018autoaugment}
E.~D. Cubuk, B.~Zoph, D.~Mane, V.~Vasudevan, and Q.~V. Le.
\newblock Auto{A}ugment: {L}earning {A}ugmentation {P}olicies from {D}ata.
\newblock {\em arXiv preprint arXiv:1805.09501}, 2018.
\newblock [Online]. Available: \url{https://arxiv.org/abs/1805.09501}.

\bibitem{2017cutout}
T.~{DeVries} and G.~W. {Taylor}.
\newblock {Improved Regularization of Convolutional Neural Networks with
  Cutout}.
\newblock {\em ArXiv e-prints}, Aug. 2017.
\newblock [Online]. Available: \url{https://arxiv.org/abs/1708.04552}.

\bibitem{2018nassurvey}
T.~{Elsken}, J.~{Hendrik Metzen}, and F.~{Hutter}.
\newblock {Neural Architecture Search: A Survey}.
\newblock {\em ArXiv e-prints}, Aug. 2018.
\newblock [Online]. Available: \url{http://arxiv.org/abs/1808.05377}.

\bibitem{gastaldi2017shakeshake}
X.~Gastaldi.
\newblock Shake-{S}hake regularization.
\newblock In {\em International Conference on Learning Representations}, 2017.

\bibitem{guo2018network}
J.~Guo, Y.~Li, W.~Lin, Y.~Chen, and J.~Li.
\newblock Network decoupling: From regular to depthwise separable convolutions.
\newblock {\em arXiv preprint arXiv:1808.05517}, 2018.
\newblock [Online]. Available: \url{https://arxiv.org/abs/1808.05517}.

\bibitem{Han2017PyramidNet}
D.~Han, J.~Kim, and J.~Kim.
\newblock Deep {P}yramidal {R}esidual {N}etworks.
\newblock {\em 2017 IEEE Conference on Computer Vision and Pattern Recognition
  (CVPR)}, Jul 2017.

\bibitem{he2015resnet}
K.~{He}, X.~{Zhang}, S.~{Ren}, and J.~Sun.
\newblock Deep {R}esidual {L}earning for {I}mage {R}ecognition.
\newblock In {\em IEEE Conference on Computer Vision and Pattern Recognition
  (CVPR)}, pages 770--778, 2016.

\bibitem{huang2018gpipe}
Y.~Huang, Y.~Cheng, D.~Chen, H.~Lee, J.~Ngiam, Q.~V. Le, and Z.~Chen.
\newblock Gpipe: {E}fficient {T}raining of {G}iant {N}eural {N}etworks using
  {P}ipeline {P}arallelism.
\newblock {\em arXiv preprint arXiv:1811.06965}, 2018.
\newblock [Online]. Available: \url{https://arxiv.org/abs/1811.06965}.

\bibitem{hundt2018hypertree}
A.~Hundt, V.~Jain, C.~Paxton, and G.~D. Hager.
\newblock {The CoSTAR Block Stacking Dataset: Learning with Workspace
  Constraints}.
\newblock {\em ArXiv e-prints}, Oct. 2018.
\newblock [Online]. Available: \url{https://arxiv.org/abs/1810.11714}.

\bibitem{liu2018PNAS}
C.~Liu, B.~Zoph, M.~Neumann, J.~Shlens, W.~Hua, L.~Li, L.~Fei{-}Fei, A.~L.
  Yuille, J.~Huang, and K.~Murphy.
\newblock Progressive neural architecture search.
\newblock In {\em European Conference on Computer Vision}, 2018.

\bibitem{liu2018darts}
H.~Liu, K.~Simonyan, and Y.~Yang.
\newblock {DARTS}: {D}ifferentiable {A}rchitecture {S}earch.
\newblock In {\em International Conference on Learning Representations}, 2019.

\bibitem{loshchilov2016sgdrcosineannealing}
I.~Loshchilov and F.~Hutter.
\newblock Sgdr: Stochastic gradient descent with warm restarts.
\newblock In {\em International Conference on Learning Representations}, 2017.

\bibitem{pmlr-v64-mendoza_towards_2016}
H.~Mendoza, A.~Klein, M.~Feurer, J.~T. Springenberg, and F.~Hutter.
\newblock Towards automatically-tuned neural networks.
\newblock In F.~Hutter, L.~Kotthoff, and J.~Vanschoren, editors, {\em
  Proceedings of the Workshop on Automatic Machine Learning}, volume~64 of {\em
  Proceedings of Machine Learning Research}, pages 58--65, New York, New York,
  USA, 24 Jun 2016. PMLR.

\bibitem{orlin1997polynomialnetworksimplex}
J.~B. Orlin.
\newblock A polynomial time primal network simplex algorithm for minimum cost
  flows.
\newblock {\em Mathematical Programming}, 78(2):109--129, 1997.

\bibitem{paszke2017pytorch}
A.~Paszke, S.~Gross, S.~Chintala, G.~Chanan, E.~Yang, Z.~DeVito, Z.~Lin,
  A.~Desmaison, L.~Antiga, and A.~Lerer.
\newblock Automatic differentiation in {P}y{T}orch.
\newblock In {\em NeurIPS-W}, 2017.

\bibitem{2018enas}
H.~{Pham}, M.~Y. {Guan}, B.~{Zoph}, Q.~V. {Le}, and J.~{Dean}.
\newblock Efficient {N}eural {A}rchitecture {S}earch via {P}arameters
  {S}haring.
\newblock {\em International Conference on Machine Learning}, pages 4092--4101,
  2018.

\bibitem{real2018regularized}
E.~Real, A.~Aggarwal, Y.~Huang, and Q.~V. Le.
\newblock Regularized {E}volution for {I}mage {C}lassifier {A}rchitecture
  {S}earch.
\newblock {\em arXiv preprint arXiv:1802.01548}, 2018.
\newblock [Online]. Available: \url{https://arxiv.org/abs/1802.01548}.

\bibitem{real2017large}
E.~Real, S.~Moore, A.~Selle, S.~Saxena, Y.~L. Suematsu, J.~Tan, Q.~V. Le, and
  A.~Kurakin.
\newblock Large-scale evolution of image classifiers.
\newblock In {\em Proceedings of the 34th International Conference on Machine
  Learning-Volume 70}, pages 2902--2911. JMLR. org, 2017.

\bibitem{recht2018cifar10}
B.~Recht, R.~Roelofs, L.~Schmidt, and V.~Shankar.
\newblock Do {CIFAR}-10 {C}lassifiers {G}eneralize to {CIFAR}-10?
\newblock {\em arXiv preprint arXiv:1806.00451}, 2018.
\newblock [Online]. Available: \url{https://arxiv.org/abs/1806.00451}.

\bibitem{s2018mobilenetv2}
M.~Sandler, A.~Howard, M.~Zhu, A.~Zhmoginov, and L.-C. Chen.
\newblock Mobile{N}et{V}2: {I}nverted {R}esiduals and {L}inear {B}ottlenecks.
\newblock {\em CVPR}, 2018.

\bibitem{NIPS2016_ConvNeuralFabrics}
S.~Saxena and J.~Verbeek.
\newblock Convolutional neural fabrics.
\newblock In D.~D. Lee, M.~Sugiyama, U.~V. Luxburg, I.~Guyon, and R.~Garnett,
  editors, {\em Advances in Neural Information Processing Systems 29}, pages
  4053--4061. Curran Associates, Inc., 2016.

\bibitem{szegedy2016rethinking_incptionv3}
C.~{Szegedy}, V.~{Vanhoucke}, S.~{Ioffe}, J.~{Shlens}, and Z.~{Wojna}.
\newblock Rethinking the {I}nception {A}rchitecture for {C}omputer {V}ision.
\newblock In {\em IEEE Conference on Computer Vision and Pattern Recognition
  (CVPR)}, pages 2818--2826, 2016.

\bibitem{zoph2017neural}
B.~{Zoph} and Q.~V. {Le}.
\newblock Neural architecture search with reinforcement learning.
\newblock {\em International Conference on Learning Representations}, 2017.

\bibitem{2017nasnet}
B.~{Zoph}, V.~{Vasudevan}, J.~{Shlens}, and Q.~V. {Le}.
\newblock Learning transferable architectures for scalable image recognition.
\newblock {\em Computer Vision and Pattern Recognition (CVPR)}, 2018.

\end{thebibliography}
}

\end{document}